\documentclass[a4paper,fleqn]{cas-dc}

\usepackage[authoryear,longnamesfirst]{natbib}
\usepackage{booktabs}
\usepackage{tabularx}
\usepackage{multirow}
\usepackage{amsmath}
\usepackage{graphicx}

\newif\ifanonym
\anonymfalse

\begin{document}
\let\WriteBookmarks\relax
\def\floatpagepagefraction{1}
\def\textpagefraction{.001}

\shorttitle{A Comparative Explainability Framework for DeBERTa-v3}

\ifanonym
  \shortauthors{Anonymous Author(s)}
\else
  \shortauthors{J. D\'iaz et al.}
\fi

\title[mode=title]{A Comparative Explainability Framework for DeBERTa-v3 in Zero-Shot Medical Abstract Classification}

\ifanonym
  % ----- Anonymized block (peer-review version) -----
  \author[1]{Anonymous Author 1}
  \cormark[1]
  \affiliation[1]{organization={Anonymous Institution},
              addressline={},
              city={},
              postcode={},
              state={},
              country={}}
  \cortext[1]{Corresponding author}
\else
  % ===== TODO: Author metadata (fill in before submission) =====
  % Javier: \ead{<email>}, \credit{Writing - Original Draft, Conceptualization, Methodology, Software, Investigation}
  % Juan:   \ead{<email>}, \credit{Writing - Review \& Editing, Supervision}
  % David:  \ead{<email>}, \credit{Writing - Review \& Editing, Supervision, Funding acquisition}
  % Affiliation 1: addressline + postcode (e.g., Facultad de Informática, Campus de Espinardo; 30100)
  % Then remove this comment block and the empty \ead{}/\credit{} above.
  % ============================================================
\author[1]{Javier D\'iaz Esteban-Herreros}[orcid=0009-0007-1658-9007]
\ead{Javier.Diaz24@alu.uclm.es}
\credit{Conceptualization, Methodology, Software, Formal analysis, Investigation, Data curation, Visualization, Writing – original draft.}

\affiliation[1]{organization={Department of Technologies and Information Systems, Universidad de Castilla-La Mancha},
              addressline={Avenida Carlos III, s/n},
              city={Toledo},
              postcode={45071},
              state={},
              country={Spain}}

\author[1]{David Mu\~noz-Valero}[orcid=0000-0002-2509-9911]
\ead{david.munoz@uclm.es}
\credit{Conceptualization, Methodology, Validation, Writing – original draft, Writing – review \& editing, Supervision.}

\author[2]{Raquel Mart\'inez-Espa\~na}[orcid=0000-0002-6750-2203]
\ead{raquel.m.e@um.es}
\credit{Conceptualization, Methodology, Validation, Writing – review \& editing, Supervision.}

\author[2,3]{Jose M. Juarez}[orcid=0000-0003-1776-1992]
\ead{jmjuarez@um.es}
\credit{Conceptualization, Methodology, Resources, Writing – review \ editing, Supervision.}

\affiliation[2]{organization={University of Murcia},
              addressline={Avenida Teniente Flomesta, 5 },
              city={Murcia},
              postcode={3003},
              country={Spain}}
\affiliation[3]{organization={Murcian Bio-Health Institute (IMIB-Arrixaca)},
              addressline={Pabellón Docente del Hospital Clínico Universitario Virgen de la Arrixaca},
              city={Murcia},
              postcode={3120},
              country={Spain}}

\author[1]{Juan Moreno-Garc\'ia}[orcid=0000-0003-2430-145X]
\cormark[1]
\ead{juan.moreno@uclm.es}
\credit{Conceptualization, Methodology, Resources, Writing – review \& editing, Supervision.}

\cortext[1]{Corresponding author}
\fi

\begin{abstract}
A comparative explainability framework is presented to audit DeBERTa-v3 under zero-shot classification of medical abstracts. The work addresses the disagreement problem in Explainable Artificial Intelligence, where different attribution methods produce divergent explanations for the same input and prediction. A natural language inference engine is implemented over the Medical Abstracts corpus with five enriched hypotheses per diagnostic category and a balanced sample of one thousand texts per class. Five explanation methods are compared: SHAP and LIME as model-agnostic approaches, occlusion and Input x Gradient as deep-learning-specific approaches, and Attention x Gradient as a transformer-specific approach. Explanations are standardized through top-token attribution, and pairwise agreement is quantified using the Jaccard index. High predictive accuracy is achieved across well-defined clinical domains, whereas performance degrades under high semantic ambiguity. Explanatory stability directly mirrors predictive certainty, exhibiting strong convergence in univalent categories and a marked drop under diagnostic uncertainty. Furthermore, qualitative error auditing uncovers three systemic failure mechanisms: lexical hypersensitivity, semantic overlap, and loss of attribution coherence. The results support the combined use of several explanation methods and quantitative agreement metrics when auditing transformer-based models in medical text classification, and suggest prioritizing specific clinical ontologies over broad diagnostic labels.

% Explanations are standardized through the twenty most relevant tokens, and pairwise agreement is quantified with the Jaccard index. Predictive accuracy exceeds 74.5\% in four specific categories, while the general pathological category drops to 6\% because of semantic overlap. Explanatory agreement spans 0.24--0.33 across the four specific categories (overall mean 0.27) and drops to 0.17 in the general category, indicating that explanation stability decreases together with predictive performance. The analysis of misclassified general cases identifies lexical hypersensitivity, polysemic overlap, and loss of attribution coherence. 
\end{abstract}

\begin{highlights}
\item Comparative explainability framework for DeBERTa-v3 medical abstract classification
\item Five explanation methods compared via Top-20 tokens and Jaccard index
\item Explanatory agreement tracks predictive performance across clinical categories
\item Generic category collapse linked to semantic overlap and lexical hypersensitivity
\item Specific clinical ontologies recommended over broad diagnostic labels
\end{highlights}


\maketitle

\section{Introduction}\label{sec:introduction}

Transformer models have become the dominant architecture in natural language processing \citep{Vaswani2017}. Their capacity to model long-range dependencies through attention mechanisms has improved many text classification tasks, including clinical text classification where transformer backbones now set the state of the art \citep{Mishra2025, Saedi2023}. However, this performance comes with an increase in architectural complexity, which turns the internal decision process into a black box that is difficult to inspect. The problem becomes more salient in clinical settings, where an automated decision may influence diagnosis and patient safety: a model that is globally accurate but locally opaque offers little reassurance to a clinician who must justify an individual prediction, and a spurious lexical cue may be confounded with a legitimate clinical descriptor \citep{EDPS2023}. The European AI Act reinforces the need for auditability in high-risk systems such as automated diagnosis \citep{DiMarino2025, Arrieta2020}, and the broader regulatory debate around the General Data Protection Regulation has further pushed the field towards post-hoc explanations that are both plausible to a human reader and technically faithful to the reasoning of the model \citep{GDPR2016, Lipton2018}.

Explainable Artificial Intelligence, or XAI, provides methods to attribute model predictions to input features \cite{dwivedi2023explainable}. These methods range from model-agnostic approaches such as SHAP \citep{Lundberg2017} and LIME \citep{lime} to model-specific approaches based on gradients or attention. A recurrent limitation reported in the literature is the disagreement problem: different explanation methods often produce inconsistent or contradictory attributions for the same prediction \citep{Longo2024, Krishna2022}. The absence of a ground-truth explanation complicates empirical validation and makes the choice of a single explainer unreliable in high-stakes domains, where a practitioner cannot easily adjudicate between two contradictory post-hoc justifications. This concern has motivated the development of standardized evaluation frameworks such as Quantus \citep{Hedstrom2023} and BEExAI \citep{Sithakoul2024}, but their adoption remains uneven, and most applied studies still rely on a single attribution method reported through visual inspection.

This paper studies that problem in a concrete medical text classification setting. DeBERTa-v3 \citep{he2021deberta} is audited under a zero-shot paradigm based on natural language inference. The model classifies medical abstracts into five diagnostic categories without task-specific training, which is a deployment scenario of practical interest in specialized clinical domains where labelled corpora are scarce. Because standard XAI libraries show limited compatibility with DeBERTa-v3, gradient and perturbation methods are implemented natively for this architecture. The framework then compares five explanation methods over the twenty most relevant tokens and quantifies pairwise agreement with the Jaccard index, so that the disagreement problem is measured directly rather than only illustrated visually. A second protocol uses SHAP to inspect misclassified cases of the general pathological category, where the model shows the lowest predictive performance, and identifies recurring failure patterns that link the predictive collapse to the structure of the explanations.

The contributions are: (i) a zero-shot inference engine with multivariate prompt engineering for medical abstract classification over DeBERTa-v3, formulated as a natural language inference problem with five enriched hypotheses per category; (ii) a comparative explainability framework that combines agnostic, deep-learning-specific, and transformer-specific methods adapted to DeBERTa-v3, filling a gap in the XAI literature where this architecture has been less studied than BERT or RoBERTa; (iii) a quantitative validation protocol based on Top-20 token attribution and the Jaccard index that measures the disagreement problem directly across method families and clinical categories; and (iv) an error analysis of the general category that documents lexical hypersensitivity, semantic overlap, and loss of attribution coherence as recurring failure patterns linking the predictive collapse to the structure of the explanations. Contribution~(i) operationalizes the zero-shot Natural Language Inference (NLI) formulation proposed by \citet{laurer_building_2023} on a medical corpus; the novelty of the paper resides in contributions~(ii)--(iv), which together expose the dependence of explanation agreement on clinical specificity and on the family of the explainer.

The remainder of the paper is organized as follows. Section~\ref{sec:related_work} reviews XAI methods for text and transformers, together with the toolkits and benchmarks that systematize their evaluation. Section~\ref{sec:methodology} details the methodology followed, including the corpus, the inference engine, the explanation methods, the standardization protocol, the disagreement quantification, and the error analysis protocol. Section~\ref{sec:experimental_setup} describes the preprocessing, reproducibility, and evaluation procedures, as well as the implementation details. Section~\ref{sec:results} reports the predictive performance, the explainability comparison, the cross-class agreement analysis, and the error analysis of the general pathological category. Section~\ref{sec:discussion} discusses the findings in the light of previous reports and of the regulatory context. Section~\ref{sec:limitations} examines the study's limitations and validity considerations. Section~\ref{sec:conclusions} concludes and outlines future work.

\section{Related Work}\label{sec:related_work}

This section reviews the background against which the present contribution is positioned. First, the work is placed within the broader taxonomy of XAI methods and the recent shift from visual to quantitative evaluation, including the regulatory pressure motivating the audit of high-stakes models. Second, attribution methods for transformer-based encoders are discussed, given that DeBERTa-v3 is the architecture under study. Finally, recent advances respecting the internal structure of transformers are summarized, the most widely used software libraries and benchmarks are contrasted through a comparative table, and the specific case of XAI applied to medical text classification is presented as the deployment domain of this paper.

% This section reviews the background against which the present contribution is positioned. We first place the work in the broader taxonomy of XAI methods and the recent shift from visual to quantitative evaluation, including the regulatory pressure that motivates the audit of high-stakes models. We then discuss attribution methods for transformer-based encoders, since DeBERTa-v3 is the architecture under study. We summarize recent advances that respect the internal structure of transformers, we contrast the most widely used software libraries and benchmarks through a comparative table, and we close with the specific case of XAI applied to medical text classification, which is the deployment domain of this paper.

\subsection{XAI taxonomies and evaluation}\label{subsec:xai_taxonomies}

XAI methods can be classified by the moment of explanation generation, by their scope, and by their dependence on the model \citep{molnar, Arrieta2020}. Ante-hoc methods build interpretability into the model design, while post-hoc methods are applied after training. Model-agnostic methods operate only on inputs and outputs, whereas model-specific methods use internal parameters or gradients. Local explanations address individual predictions, and global explanations describe overall model behavior. The clinical domain requires both levels: a global audit of the system and a local justification for each decision \citep{Arrieta2020}, since a classifier that is globally accurate but locally opaque offers little reassurance to a clinician who must justify an individual prediction. The recent distinction between \emph{faithfulness} (whether the explanation reflects the actual model computation) and \emph{plausibility} (whether the explanation is convincing to a human) has become central to this discussion \citep{Jacovi2020, Lipton2018}, and the cognitive notion of interpretability itself has been formalized as a multifaceted property rather than a single scalar \citep{Longo2024}.

The evaluation of explanations has moved from visual inspection to quantitative assessment. Two complementary lines of work have consolidated: sanity-check procedures, which verify that an explanation actually depends on the model rather than on artefacts of the data or the attribution code \citep{adebayo2018sanity}; and robustness analyses, which measure how stable an attribution is under small perturbations of the input or the model parameters \citep{AlvarezMelis2018}. Frameworks such as Quantus \citep{Hedstrom2023} and BEExAI \citep{Sithakoul2024} provide standardized metrics for fidelity, robustness, and complexity, and recent work attempts a unifying categorical foundation that organizes the proliferation of metrics into a coherent taxonomy \citep{Giannini2024}. Despite these advances, the disagreement problem remains a central concern: when explanations are inconsistent, their practical use becomes difficult to justify \citep{Longo2024, Krishna2022}. This situation is especially delicate in medical applications, where spurious correlations can be confused with legitimate clinical features \citep{EDPS2023}, and where a regulatory push toward auditability has emerged in the European AI Act \citep{DiMarino2025} and the broader debate on explainability requirements under the General Data Protection Regulation \citep{GDPR2016}.

\subsection{XAI for transformer-based models}\label{subsec:xai_transformers}

Early work questioned the interpretability of attention weights. \citet{serrano2019is} showed that attention is not always a reliable explanation, because attention heads can highlight syntactically necessary tokens without causal influence on the prediction. \citet{jain2019attention} reported similar evidence and motivated hybrid approaches that combine attention with gradient information. \citet{chefer2021transformer} proposed a transformer interpretability method based on conservative propagation, which has influenced subsequent attribution techniques for attention-based models. A complementary line of work applies attention flow to quantify how information propagates through the stacked layers of a transformer, providing a less noisy alternative to raw attention inspection \citep{Abnar2020}, and occlusion combined with language modelling has been studied as a post-hoc variant for sequence classifiers \citep{Harbecke2020}.

Gradient-based methods such as Input x Gradient \citep{Ali2022} and integrated gradients \citep{Sundararajan2017} attribute relevance through backpropagation. The axioms of implementation invariance and completeness that motivate integrated gradients make it an appealing reference method, although its sensitivity to the choice of baseline remains an open issue. Perturbation-based methods such as occlusion \citep{RobnikSikonja2008} measure the effect of removing or replacing input tokens; their simplicity is offset by a dependence on the perturbation policy (masking token, sliding window length, replacement distribution). Model-agnostic methods such as SHAP \citep{Lundberg2017} and LIME \citep{lime,Ribeiro2016} remain widely used because they can be applied to any classifier, although they are sensitive to perturbation design and reference values, and their KernelSHAP and LIME variants incur a non-negligible variance across runs unless the perturbation budget is increased. A recent ESWA study illustrates this combined use of self-supervised transformers with LIME and SHAP to provide interpretable predictions in a high-stakes classification setting, reporting state-of-the-art accuracy together with token-level explanations \citep{Bostancioglu2026}.

\subsection{Recent advances in attribution for transformers}\label{subsec:recent_advances}

Recent methods aim to respect the internal structure of transformers. ALTI-Logit and AttnLRP decompose predictions into additive contributions across attention and feed-forward blocks \citep{arras2025quantifying, achtibat2024attnlrp}. CA-LIG integrates layer-wise gradients with class-specific attention gradients to trace how representations evolve across layers \citep{mersha2026calig}. LLiMe uses a large language model to generate semantically coherent neighborhoods for LIME, mitigating the grammatical degradation caused by token removal \citep{angiulli2025llime}. These advances are accompanied by a broader body of surveys that systematize explainability for large language models and deep models at scale \citep{Zhao2024, DiMarino2025, cheng2025comprehensive, Mehta2023, xAI2024PartIII, milcke2024}, and by applied software libraries, discussed in the next subsection, that lower the barrier to deploying these methods in practice.

\subsection{XAI in medical text classification}\label{subsec:xai_medical_text}

Explainability has been applied to text classification in medicine and in adjacent domains such as patent novelty analysis \citep{Jang2023}. In clinical text classification, explanations are expected not only to justify a label but also to indicate whether the model relies on legitimate clinical descriptors. Several ESWA contributions confirm that BERT-based architectures with attention mechanisms are now the default backbone for medical text tasks such as medical emergency prediction from unstructured clinical notes \citep{Mishra2025} and consumer health question classification \citep{Saedi2023}, although the post-hoc auditability of these models remains an open concern. Diagnostic studies of explanation techniques for text classification have shown that different methods can highlight different tokens even when the predicted label is the same \citep{Atanasova2020}; this evidence is complemented by ground-truth benchmarks such as CLEVR-XAI \citep{Arras2022}, which make it possible to evaluate attributions against a controlled reference. Faithfulness and plausibility therefore need to be distinguished explicitly \citep{Jacovi2020, Lipton2018}. This paper follows that line and evaluates several methods against each other over a shared prediction rather than against a single ground-truth explanation, which is unavailable for transformer-based NLI over free-text medical abstracts.

\subsection{Comparative XAI frameworks and toolkits}\label{subsec:toolkits}

Beyond single attribution methods, several toolkits and benchmarks have emerged to systematize the application and evaluation of XAI. Table~\ref{tab:xai_toolkits} summarizes the most relevant ones for the present study. Quantus \citep{Hedstrom2023} and BEExAI \citep{Sithakoul2024} focus on the quantitative evaluation of explanations through fidelity, robustness and complexity metrics, and are agnostic to the attribution backend. Captum \citep{Kokhlikyan2020} provides a unified PyTorch API for gradient, perturbation and occlusion methods but offers only partial support for the disentangled attention of DeBERTa-v3. Inseq \citep{Sarti2023} targets sequence generation models and exposes feature attribution utilities, while Ecco \citep{Alammar2021} and BertViz \citep{vig-2019-multiscale} emphasize visualization of attention and latent representations for encoder-only models. ALTI-Logit / AttnLRP \citep{arras2025quantifying, achtibat2024attnlrp} implement layer-wise relevance propagation tailored to transformers, and LLiMe \citep{angiulli2025llime} augments LIME with an LLM-based neighborhood generator. The table highlights that no single toolkit covers the five families of explanations compared in this paper natively for DeBERTa-v3, which motivates the native implementations detailed in Section~\ref{sec:methodology}.

\begin{table*}[!t]
\centering
\caption{Comparison of XAI frameworks and toolkits relevant to transformer attribution. \emph{Methods}: coverage of attribution families. \emph{DeBERTa-v3}: native support for disentangled attention. \emph{Eval metrics}: built-in quantitative evaluation.}
\label{tab:xai_toolkits}
\scriptsize
\begin{tabularx}{\textwidth}{l X l c c}
\toprule
Framework / toolkit & Primary purpose & Method families & DeBERTa-v3 & Eval metrics \\
\midrule
Quantus \citep{Hedstrom2023} & Evaluation of explanations & Agnostic backends & Partial & Yes \\
BEExAI \citep{Sithakoul2024} & Benchmark of XAI quality & Agnostic backends & Partial & Yes \\
Captum \citep{Kokhlikyan2020} & PyTorch attribution API & Gradient, occlusion, LIME, SHAP & Partial & No \\
Inseq \citep{Sarti2023} & Sequence-model interpretability & Gradient, attention, perturbation & Partial & Basic \\
Ecco \citep{Alammar2021} & Visualization of transformers & Attention, gradient & No & No \\
BertViz \citep{vig-2019-multiscale} & Attention visualization & Attention & No & No \\
ALTI-Logit / AttnLRP \citep{achtibat2024attnlrp} & LRP for transformers & Layer-wise relevance & Limited & Basic \\
LLiMe \citep{angiulli2025llime} & LLM-augmented LIME & Perturbation (LIME) & Yes (model-agnostic) & No \\
CA-LIG \citep{mersha2026calig} & Layer-wise integrated gradients & Gradient & Limited & No \\
\bottomrule
\end{tabularx}
\end{table*}

\subsection{Background: DeBERTa-v3 and zero-shot NLI}\label{subsec:background}

DeBERTa-v3 \citep{he2021deberta} improves on the original BERT architecture through three main innovations: disentangled attention, which represents each token as two vectors (content and position) and computes their interaction through a decomposable attention score; an enhanced mask decoder that predicts masked tokens only at the output of the decoder rather than at every layer, which mitigates the pre-training mismatch that arises when attention is computed on hidden states that already carry mask information; and a scale-invariant variant of the ELECTRA-style replaced-token-detection objective, which replaces the BERT-style magnitude invariance by a scale-invariant formulation that removes the dependence on the magnitude of the hidden representations. Together, these innovations produce a more expressive pre-training signal and a sharper attention computation, which is particularly beneficial for natural language inference tasks where the entailment decision depends on fine-grained token interactions over long premises.

Zero-shot NLI leverages these representations by recasting classification as an entailment problem: the model is trained on a generic NLI corpus and, at inference time, a class label is associated with a natural-language hypothesis; the input is classified by selecting the hypothesis whose entailment probability is highest. \citet{laurer_building_2023} released \texttt{deberta-v3-large-zeroshot-v2.0}, the model used in this paper, which is obtained by fine-tuning DeBERTa-v3-large on a mixture of NLI and topic-classification datasets; the resulting model attains competitive accuracy on a wide range of topics without task-specific labels. The choice of zero-shot NLI for medical abstract classification is motivated by the scarcity of annotated clinical corpora and by the regulatory advantage of avoiding patient-level training data, but it transfers all the representational bias of the pre-training corpus to the downstream decision, which is one of the reasons that motivate the explainability audit reported here.

\subsection{Research gap addressed}\label{subsec:research_gap}

The literature review reveals three primary gaps in the explainability of clinical text classifiers. First, DeBERTa-v3 remains significantly less studied from an XAI perspective than BERT or RoBERTa, and standard toolboxes lack native support for its disentangled attention mechanism (Table\ref{tab:xai_toolkits}), requiring custom model-level implementations for gradient and perturbation methods. Second, comparative XAI studies mostly depend on qualitative visual inspection or downstream acceptability, leaving the disagreement problem unmeasured through direct quantitative metrics across explainer families. Third, existing auditing protocols evaluate predictive accuracy in isolation, without examining how attribution fragmentation and explanation instability directly account for model failure in ambiguous clinical categories. To address these limitations, this paper introduces a unified auditing pipeline that tackles each gap directly. To overcome the architectural barrier, native attribution methods spanning three algorithmic families are implemented for DeBERTa-v3\. To resolve the evaluation deficit, pairwise inter-explainer agreement is quantitatively measured using the Jaccard index over standardized Top-20 tokens. Finally, to audit model vulnerability, a SHAP-based error analysis protocol is established to diagnose the root causes of predictive collapse and attribution anomalies in semantically overlapping classes.

\section{Methodology}\label{sec:methodology}

\begin{figure*}[!t]
\centering
\includegraphics[width=1\textwidth]{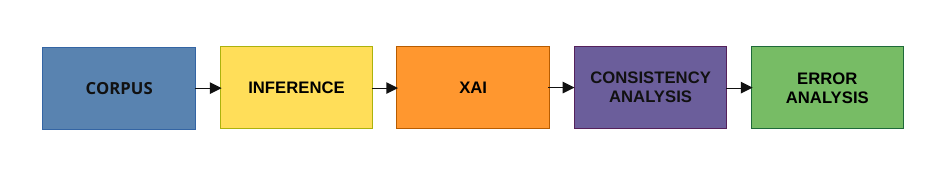}
\caption{Methodology overview. High-level workflow of the five sequential operational stages: corpus balancing, zero-shot NLI inference, XAI methods, Top-20 Jaccard consistency analysis, and SHAP-based error analysis.}
\label{fig:framework}
\end{figure*}

Figure~\ref{fig:framework} provides a high-level overview of the proposed methodological framework. To evaluate both the reliability of post-hoc explainability methods on high-certainty predictions and the root causes of predictive failures under an unsupervised zero-shot paradigm, the pipeline is structured into five sequential operational stages:

\begin{enumerate}
    \item Corpus Selection and Balancing: Establishing a balanced experimental sample from the \textit{Medical Abstracts} corpus across four specific diagnostic categories and one broad pathological class (Section \ref{subsec:corpus}).
    \item Inference Engine: Recasting multi-class clinical text classification as a NLI task using DeBERTa-v3 combined with a multivariate prompt engineering strategy based on five enriched hypotheses per category (Section \ref{subsec:inference}).
    \item Explanation Methods: Extracting local feature attributions using five distinct explanation methods spanning model-agnostic (SHAP, LIME), perturbation-based (Occlusion), gradient-based (Input $\times$ Gradient), and transformer-specific (Attention $\times$ Gradient) families (Section \ref{subsec:explanation_methods}).
    \item Consistency Analysis: Standardizing token attributions via Top-20 token selection and quantifying pairwise inter-method agreement through the Jaccard index to evaluate explanatory stability across high-certainty predictions (Section \ref{subsec:disagreement}).
    \item Error Analysis: Auditing misclassified instances within the ambiguous general pathological category using SHAP saliency maps and waterfall plots to diagnose systemic failure mechanisms (Section \ref{subsec:debugging}).
\end{enumerate}

The following sections describe each of these operational states.

\subsection{Corpus}\label{subsec:corpus}

The corpus is the Medical Abstracts dataset from \citet{Schopf2022}. The original collection contains 14,438 English medical abstracts with verified annotations. The taxonomy has five clinical categories. Four of them are specific diagnostic groups: neoplasms, digestive system diseases, nervous system diseases, and cardiovascular diseases. The fifth group, general pathological conditions, has a transversal nature and includes systemic or nonspecific states such as fever, fatigue, or diffuse inflammation. This last category introduces semantic overlap with the specific groups and motivates the explanation analysis.

Table~\ref{tab:corpus} reports the original distribution. To avoid prevalence bias in the explanation study, a balanced sample of 1,000 texts per category is used. Because the inference is zero-shot, this sample is drawn from the training split without data leakage for the inference itself.

\begin{table}[!t]
\centering
\caption{Original distribution of the Medical Abstracts corpus by clinical category.}
\label{tab:corpus}
\begin{tabular}{lrrr}
\toprule
Category & Train & Test & Total \\
\midrule
Neoplasms & 2,530 & 633 & 3,163 \\
Digestive & 1,195 & 299 & 1,494 \\
Nervous & 1,540 & 385 & 1,925 \\
Cardiovascular & 2,441 & 610 & 3,051 \\
General & 3,844 & 961 & 4,805 \\
\midrule
Total & 11,550 & 2,888 & 14,438 \\
\bottomrule
\end{tabular}
\end{table}

\subsection{Inference Engine}\label{subsec:inference}

The inference engine formulates classification as a natural language inference problem. Each medical abstract is the premise, and the category descriptions are the hypotheses. Given a premise $x$ and a candidate hypothesis $h_c$ for category $c \in \{1,\dots,C\}$, the DeBERTa-v3 encoder produces three logits $(\ell_{\text{entail}}, \ell_{\text{neutral}}, \ell_{\text{contrad}})$ that score the compatibility between $x$ and $h_c$ \citep{laurer_building_2023}. A softmax over the entailment logit and its alternatives converts this vector into an entailment probability $p_c = \sigma(\ell_{\text{entail}};\, \ell_{\text{neutral}}, \ell_{\text{contrad}})$ quantifying the degree to which the premise entails that the abstract belongs to category $c$.

The DeBERTa-v3 model \citep{laurer_building_2023, he2021deberta} used is \texttt{MoritzLaurer/deberta-v3-large-zeroshot-v2.0}.

\begin{table*}[!t]
\centering
\caption{Enriched hypotheses per clinical category used by the natural language inference engine.}
\label{tab:hypotheses}
\scriptsize
\begin{tabularx}{\textwidth}{l X}
\toprule
Category & Selected hypotheses \\
\midrule
Neoplasms & Description of a neoplastic condition such as tumor, cancer, malignancy, carcinoma, sarcoma, lymphoma, leukemic process, or metastatic disease; abnormal cell growth consistent with an oncological process; diagnosis, progression, treatment, staging, or complications of a neoplastic disease; oncological process affecting tissues or organs, including primary or metastatic tumors; pathology related to cancer or tumor biology. \\
Digestive & Disorder of the digestive system affecting the gastrointestinal tract, liver, pancreas, or biliary system; gastrointestinal symptoms, inflammation, infection, obstruction, hepatobiliary disease, or pancreatic pathology; diseases such as gastritis, colitis, pancreatitis, hepatitis, cirrhosis, intestinal dysfunction, or gastrointestinal bleeding; gastroenterological pathology involving digestive organs; digestive disease affecting esophagus, stomach, intestines, colon, liver, or pancreas. \\
Nervous & Neurological disorder affecting the brain, spinal cord, or peripheral nerves; stroke, neuropathy, seizures, demyelination, meningitis, encephalopathy, or neurodegenerative diseases; neurological deficits, altered consciousness, seizures, focal findings, or neuroinflammatory pathology; pathology affecting neural tissue or circuits; diseases of the central or peripheral nervous system. \\
Cardiovascular & Cardiovascular disorder affecting the heart or blood vessels; myocardial infarction, heart failure, arrhythmia, hypertension, ischemia, or vascular pathology; cardiological symptoms, hemodynamic changes, perfusion abnormalities, or vascular obstruction; diseases of the heart muscle, coronary arteries, valves, or systemic circulation; cardiovascular pathology affecting cardiac or vascular structures. \\
General & Systemic or generalized medical condition affecting multiple organs; generalized pathological state with symptoms such as fever, malaise, fatigue, weakness, or systemic inflammation; disorder affecting the body globally rather than a single organ system; diffuse or nonspecific organ anomaly involving general physiological dysfunction; systemic medical problem without a clearly dominant organ system. \\
\bottomrule
\end{tabularx}
\end{table*}

To reduce token sensitivity, five enriched hypotheses $\{h_c^{(1)},\dots,h_c^{(5)}\}$ are defined per category $c$; Table~\ref{tab:hypotheses} summarizes them. For each premise, the five entailment probabilities of a category are averaged
\begin{equation}
\bar{p}_c = \frac{1}{5}\sum_{r=1}^{5} p_c^{(r)},
\label{eq:aggregate}
\end{equation}
yielding an aggregated score per category. A softmax across categories then produces a normalized confidence distribution
\begin{equation}
s_c = \frac{\mathrm{e}^{\bar{p}_c}}{\sum_{c'=1}^{C}\mathrm{e}^{\bar{p}_{c'}}},
\label{eq:softmax}
\end{equation}
and the final label is assigned by the maximum confidence value $\hat{c} = \arg\max_c s_c$. Operating under zero-shot learning means that no supervised training stage is performed on the target corpus: the model is used as released by its pretraining, and the only task-specific intervention is the design of the hypotheses, which acts as a form of multivariate prompt engineering. This choice eliminates the risk of data leakage between the training split and the evaluation, but transfers all the representational bias of the pretraining corpus to the downstream decision, an aspect revisited in Section~\ref{sec:limitations}.

\subsection{Explanation Methods}\label{subsec:explanation_methods}

Five explanation methods are selected. Two requirements guide the selection: class specificity, so that the explanation is computed for the predicted category rather than for a generic model output, and preservation of polarity, so that positive and negative contributions can be distinguished and the direction of the token's influence on the predicted class can be inspected. The five methods belong to three families with different mathematical foundations: model-agnostic perturbation methods (SHAP, LIME), a local perturbation method (occlusion), a gradient method (Input x Gradient), and a transformer-specific method (Attention x Gradient). Computing a method from each family makes it possible to expose cross-family divergence, which is the empirical signature of the disagreement problem \citep{Krishna2022}.

\paragraph{Model-agnostic methods.} SHAP \citep{Lundberg2017} and LIME \citep{lime, Ribeiro2016} are applied through their Python libraries. Both methods are configured with 1,000 perturbations per sample to stabilize the attributions, which is well above the default budget of the libraries and is required to keep the estimator variance within a tolerable band over long medical abstracts. SHAP estimates the Shapley value of each token by averaging its marginal contribution across coalitions of features, producing an additive decomposition of the difference between the expected model output and the actual prediction; LIME fits a locally linear surrogate on a neighborhood of perturbed inputs and reads the coefficients of the surrogate as token attributions. Both methods return signed attributions that can be compared to the gradient- and occlusion-based outputs. The generative variant LLiMe \citep{angiulli2025llime} is not included because of its computational cost for this volume of data, which would require the invocation of a large language model to generate semantically coherent perturbations for each of the 5,000 abstracts.

\paragraph{Perturbation method.} Occlusion is implemented as a local perturbation over the premise tokens. Each token between the special markers is replaced by a neutral token, and the attribution is the difference between the original entailment probability and the perturbed probability \citep{RobnikSikonja2008}:
\begin{equation}
A_i = f_c(x) - f_c(x_{\text{mask}_i}).
\label{eq:occlusion}
\end{equation}

where $i$ denotes the positional index of the evaluated token within the input text $x$, $f_c(x)$ represents the model's output entailment probability for the target class $c$ given the complete input, and $x_{\text{mask}_i}$ is the modified input sequence where the token at position $i$ has been replaced by the neutral \texttt{[MASK]} token.

\paragraph{Gradient method.} Input x Gradient computes token-level importance by multiplying the embedding activations by the gradient of the target class score with respect to the input \cite{Ali2022}. To reduce the $d$-dimensional embedding vector to a single scalar attribution $R(x_i)$ per token while preserving sign polarity, the summation across all $d$ embedding dimensions is explicitly incorporated into the formula rather than applying an $L_2$ norm:
\begin{equation}
R(x_i) = \sum_{k=1}^{d} \left( x_{i,k} \cdot \frac{\partial y_c}{\partial x_{i,k}} \right)
\label{eq:ixg}
\end{equation}
% The embedding vector is reduced to a scalar per token through scalar summation rather than the L2 norm. This choice preserves the sign of the contribution and keeps the explanation comparable with SHAP, LIME, and occlusion. Integrated gradients are not used because preliminary tests showed unstable attributions over subword fragments and higher latency.

where $i$ identifies the token index in the input sequence, $c$ represents the target diagnostic category under evaluation, $x_{i,k}$ is the $k$-th element of the $d$-dimensional embedding vector $x_i \in \mathbb{R}^d$, $y_c$ is the logit score assigned by the model to class $c$, and $\frac{\partial y_c}{\partial x_{i,k}}$ denotes the partial derivative of $y_c$ with respect to $x_{i,k}$ computed via backpropagation.

\paragraph{Transformer-specific method.} Attention x Gradient combines attention weights with the gradient of the prediction with respect to those weights \citep{chefer2021transformer}:
\begin{equation}
\text{Importance}_{i,c} = \alpha_i \times \frac{\partial y_c}{\partial \alpha_i}.
\label{eq:axg}
\end{equation}

where $i$ specifies the token position, $c$ is the target diagnostic class, $\alpha_i$ represents the scalar attention weight directed from the special classification token \texttt{[CLS]} which aggregates the sequence-level semantic representation used for the final prediction to token $i$ (extracted from the final encoder layer and averaged across all attention heads), and $\frac{\partial y_c}{\partial \alpha_i}$ is the partial derivative of the class score $y_c$ with respect to $\alpha_i$. This formulation mitigates the noise reported for raw attention weights \citep{serrano2019is} by combining mechanical information flow ($\alpha_i$) with predictive causal sensitivity ($\frac{\partial y_c}{\partial \alpha\_i}$), ensuring that attributions are class-specific to category $c$ and retain sign polarity.

% \subsection{Standardization protocol}\label{subsec:standardization}

% To compare methods with different mathematical bases, all attributions are standardized. For each explanation produced by one of the five methods, the $k=20$ most relevant tokens are selected by absolute attribution value, and the polarity is retained only for visualization. The threshold $k=20$ follows the range suggested by \citet{chefer2021transformer} and is intended to capture the main clinical entities of a medical abstract while limiting statistical noise from secondary tokens. Preliminary inspection showed that smaller thresholds ($k \in \{5,10\}$) over-emphasized a few dominant descriptors and increased the variance of the Jaccard index across pairs, whereas larger thresholds ($k \in \{50,100\}$) diluted the clinical core with particles and stopwords without substantially changing the comparative pattern; the choice of $k=20$ therefore reflects a trade-off between signal and noise that is consistent with the median length of the diagnostic phrases in the corpus.

\subsection{Consistency Analysis}\label{subsec:disagreement}

% Pairwise agreement between explanation methods is quantified with the Jaccard index over the Top-20 token sets \citep{Krishna2022}:
% \begin{equation}
% J(A,B) = \frac{|A \cap B|}{|A \cup B|},
% \label{eq:jaccard}
% \end{equation}
% where $A$ and $B$ are the Top-20 token sets produced by two explanation methods for the same predicted case. A value close to one indicates that two methods agree on the relevant tokens, while a value close to zero indicates divergence. The index is computed for every pair of methods and for each clinical category.

% The Jaccard index is adopted over alternative set-overlap measures for three reasons. First, it is threshold-free once the Top-$k$ sets are defined, which removes the need for an additional hyper-parameter. Second, it weighs intersection and union symmetrically, so that two methods each highlighting a different small set of tokens receive a low score even if the union is large; this property penalizes the kind of divergence that characterizes the disagreement problem. Third, the index is the only quantitative metric used in the comparative phase of this paper, as recommended by \citet{Krishna2022} for direct disagreement assessment; complementary ranking-aware measures such as the Rank-Biased Overlap \citep{AlvarezMelis2018} or fidelity metrics such as AOPC-sufficiency and comprehensiveness \citep{Hedstrom2023} are explicitly identified in Section~\ref{sec:limitations} as future work, since their computation requires an additional experimental protocol that lies outside the scope of the present study.

To address the disagreement problem across XAI frameworks with divergent mathematical foundations, a unified consistency analysis protocol is established, integrating attribution standardization and pairwise agreement quantification.

Attribution Standardization (Top-20 Tokens): To compare methods with different mathematical bases, all attributions are standardized. For each explanation produced by one of the five methods, the $k=20$ most relevant tokens are selected by absolute attribution value, and the polarity is retained only for visualization. The threshold $k=20$ follows the range suggested by \citet{chefer2021transformer} and is intended to capture the main clinical entities of a medical abstract while limiting statistical noise from secondary tokens. Preliminary inspection showed that smaller thresholds ($k \in \{5,10\}$) over-emphasized a few dominant descriptors and increased the variance of the Jaccard index across pairs, whereas larger thresholds ($k \in \{50,100\}$) diluted the clinical core with particles and stopwords without substantially changing the comparative pattern; the choice of $k=20$ therefore reflects a trade-off between signal and noise that is consistent with the median length of the diagnostic phrases in the corpus.

Pairwise Agreement Quantification (Jaccard Index): Pairwise agreement between explanation methods is quantified with the Jaccard index over the Top-20 token sets \citep{Krishna2022}:
\begin{equation}
J(A,B) = \frac{|A \cap B|}{|A \cup B|},
\label{eq:jaccard}
\end{equation} where $A$ and $B$ are the Top-20 token sets produced by two explanation methods for the same predicted case. A value close to one indicates that two methods agree on the relevant tokens, while a value close to zero indicates divergence. The index is computed for every pair of methods and for each clinical category.

The Jaccard index is adopted over alternative set-overlap measures for three reasons. First, it is threshold-free once the Top-$k$ sets are defined, which removes the need for an additional hyper-parameter. Second, it weighs intersection and union symmetrically, so that two methods each highlighting a different small set of tokens receive a low score even if the union is large; this property penalizes the kind of divergence that characterizes the disagreement problem. Third, the index is the only quantitative metric used in the comparative phase of this paper, as recommended by \citet{Krishna2022} for direct disagreement assessment; complementary ranking-aware measures such as the Rank-Biased Overlap \citep{AlvarezMelis2018} or fidelity metrics such as AOPC-sufficiency and comprehensiveness \citep{Hedstrom2023} are explicitly identified in Section~\ref{sec:limitations} as future work, since their computation requires an additional experimental protocol that lies outside the scope of the present study.

\subsection{Error Analysis}\label{subsec:debugging}

A second protocol inspects the general pathological category, where the model shows the lowest accuracy. SHAP is used as a error analysis tool because it combines class specificity, polarity preservation, and a principled Shapley-value foundation that is robust to correlated features, which makes it suitable for interpreting individual predictions over long medical abstracts. For each misclassified case, the twenty most relevant tokens are visualized through saliency maps and waterfall plots. Saliency maps show positive contributions in magenta and negative contributions in blue. Waterfall plots quantify the marginal contribution of each token to the final prediction, decomposing the gap between the expected model output and the actual prediction into additive token contributions. The same 1,000-perturbation budget used in the comparative phase is retained here to ensure that the visualization is statistically stable.

\section{Experimental Setup}\label{sec:experimental_setup}

This section describes the technical conditions under which the methodology of Section~\ref{sec:methodology} is executed. We first report the software stack and the hardware used for the GPU-intensive stages (Section \ref{subsec:implementation}); we then describe the preprocessing applied to the Medical Abstracts corpus before sampling (Section \ref{subsec:preprocessing}); we state the reproducibility measures adopted to make the pipeline inspectable (Section \ref{subsec:reproducibility}); and we define the evaluation metrics used to report predictive performance and explanation agreement (Section \ref{subsec:metrics}).

\subsection{Implementation details}\label{subsec:implementation}

The implementation is developed in Python 3.10.13 over Jupyter notebooks, which provides a cohesive workflow for executing the deep-learning inference, the statistical analysis of the clinical metrics, and the real-time documentation of the findings. The model is loaded through the Hugging Face Transformers library \citep{wolf-etal-2020-transformers}. Data handling uses Pandas \citep{reback2020pandas} and NumPy \citep{harris2020array}. Evaluation metrics are computed with Scikit-learn \citep{scikit-learn}. Visualizations use Matplotlib \citep{Hunter:2007} and Seaborn \citep{Waskom2021}. SHAP \citep{Lundberg2017} and LIME \citep{lime} are used for the agnostic explanations, while occlusion, Input x Gradient, and Attention x Gradient are implemented natively for DeBERTa-v3. The choice of native implementations is forced by the limited compatibility of standard XAI libraries with the disentangled attention mechanism of DeBERTa-v3, as discussed in Section~\ref{subsec:toolkits}.

The model has approximately 400 million parameters and is deployed on an Nvidia RTX 4050 GPU. The high computational demand of transformer inference at this scale, combined with the intensive gradient computation required by the attribution methods, is the main factor that bounds the evaluation to the 1{,}000-texts-per-category sample. A fixed random seed is used throughout the pipeline to reduce the variance of stochastic operations and to support reproducibility, mitigating the inherent stochasticity of the neural network and ensuring that the fidelity audits are consistent across successive runs of the pipeline.

\subsection{Preprocessing}\label{subsec:preprocessing}

The balanced sample is obtained through random sampling of 1{,}000 texts per category from the training split of the Medical Abstracts corpus, which contains 11{,}550 abstracts across the five categories. Sampling from the training split is admissible because the inference is zero-shot and therefore no supervised training is performed on the target corpus, which eliminates the risk of data leakage between the training and evaluation phases. Once sampled, the abstracts are loaded into a Pandas DataFrame and the original integer labels in the range 1--5 are converted to the zero-based range 0--4 required by the Hugging Face interface. No stemming, lemmatization, stopword removal, or other linguistic normalization is applied, because the transformer tokenizer operates directly on the raw text and preserves the lexical information that the explanation methods attribute. The hypotheses described in Table~\ref{tab:hypotheses} are prepended to each abstract following the natural language inference template of the model, and the resulting premise--hypothesis pairs are tokenized with the DeBERTa-v3 tokenizer before being fed to the inference engine described in Section~\ref{subsec:inference}.

\subsection{Reproducibility}\label{subsec:reproducibility}

The inference engine, the explanation methods, and the validation protocol are implemented as separate stages so that the outputs of each stage can be inspected and reused independently. The intermediate artefacts of the pipeline (the per-abstract entailment probabilities, the per-token attributions of each method, and the per-pair Jaccard values) are persisted to disk so that the comparative analysis can be reproduced without re-running the GPU-intensive stages. The 1{,}000-perturbation budget used for SHAP and LIME is held fixed across all samples to ensure that the estimator variance is comparable across methods and categories, and the same budget is retained for the SHAP-based error analysis phase.

\subsection{Evaluation metrics}\label{subsec:metrics}

Predictive performance is reported through confusion matrices in both absolute counts and normalized form, per-class recall on the diagonal of the normalized matrix, and the aggregate F1 score computed by macro-averaging over the five categories. Explanation agreement is reported through pairwise Jaccard values per method pair and category, per-category averages obtained by averaging over the ten method pairs, and a global average across all method pairs and categories. The per-category standard deviations reported alongside the averages reflect the variability across method pairs, not across texts; this distinction is made explicit in Section~\ref{sec:limitations} when discussing internal validity. The error analysis phase is reported qualitatively through saliency maps and waterfall plots, complemented by the structured failure patterns summarized in tables.

\section{Experiments and Results}\label{sec:results}

This section reports the empirical evidence at three levels of analysis: the predictive performance of the zero-shot engine on the balanced sample (Section \ref{subsec:predictive_performance}), the explainability comparison across the five attribution methods (Section \ref{subsec:explainability_comparison}), and the cross-class aggregation of the agreement index that exposes the dependence of the disagreement problem on clinical specificity (Section \ref{subsec:cross_class}). A fourth subsection reports the SHAP-based error analysis of the general pathological category, the category where the model exhibits the worst predictive performance and the weakest explanation agreement (Section \ref{subsec:debugging_results}).

\begin{figure*}[!t]
\centering
\begin{minipage}{0.49\textwidth}
\centering
\includegraphics[width=\linewidth]{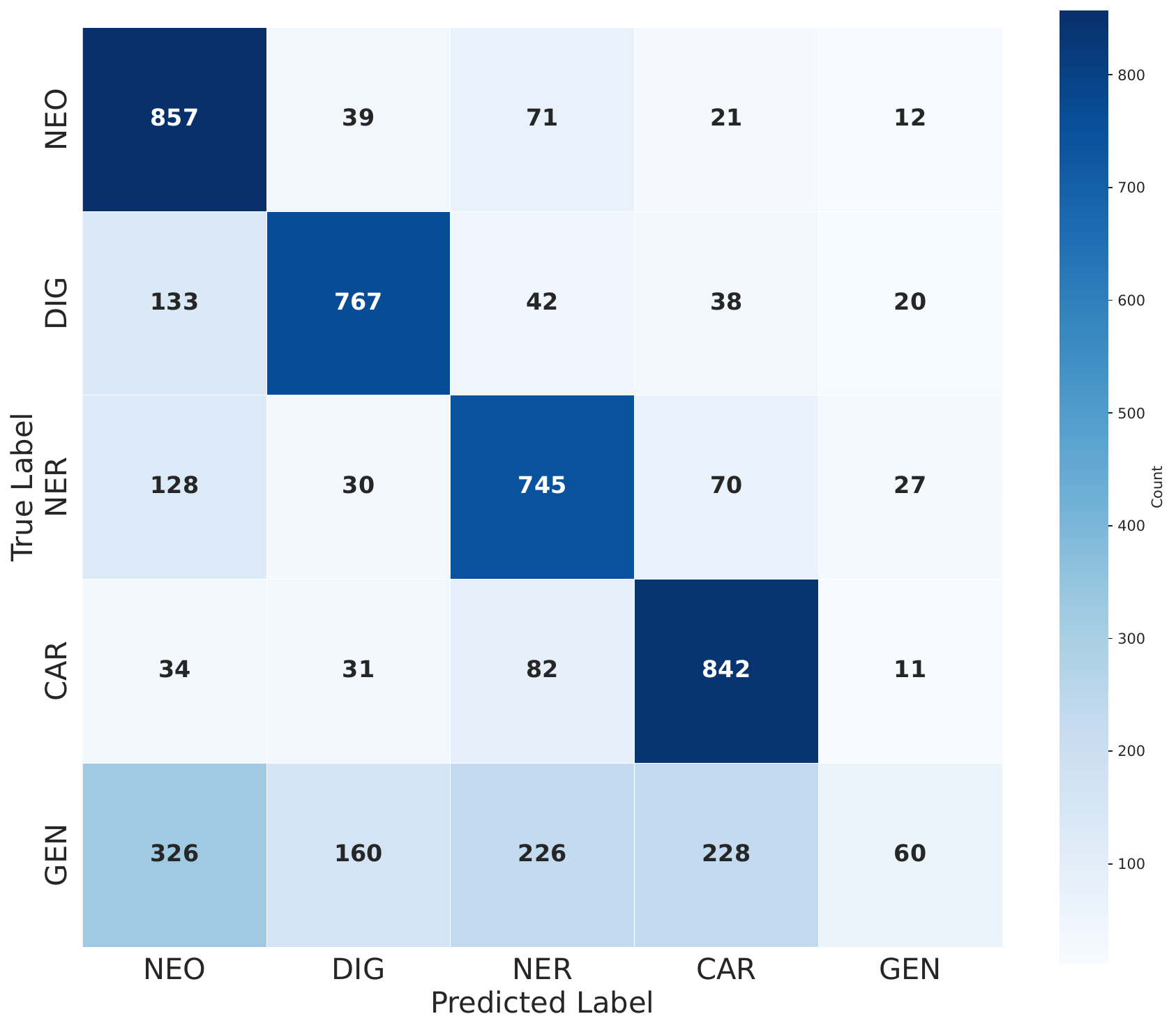}
\end{minipage}\hfill
\begin{minipage}{0.49\textwidth}
\centering
\includegraphics[width=\linewidth]{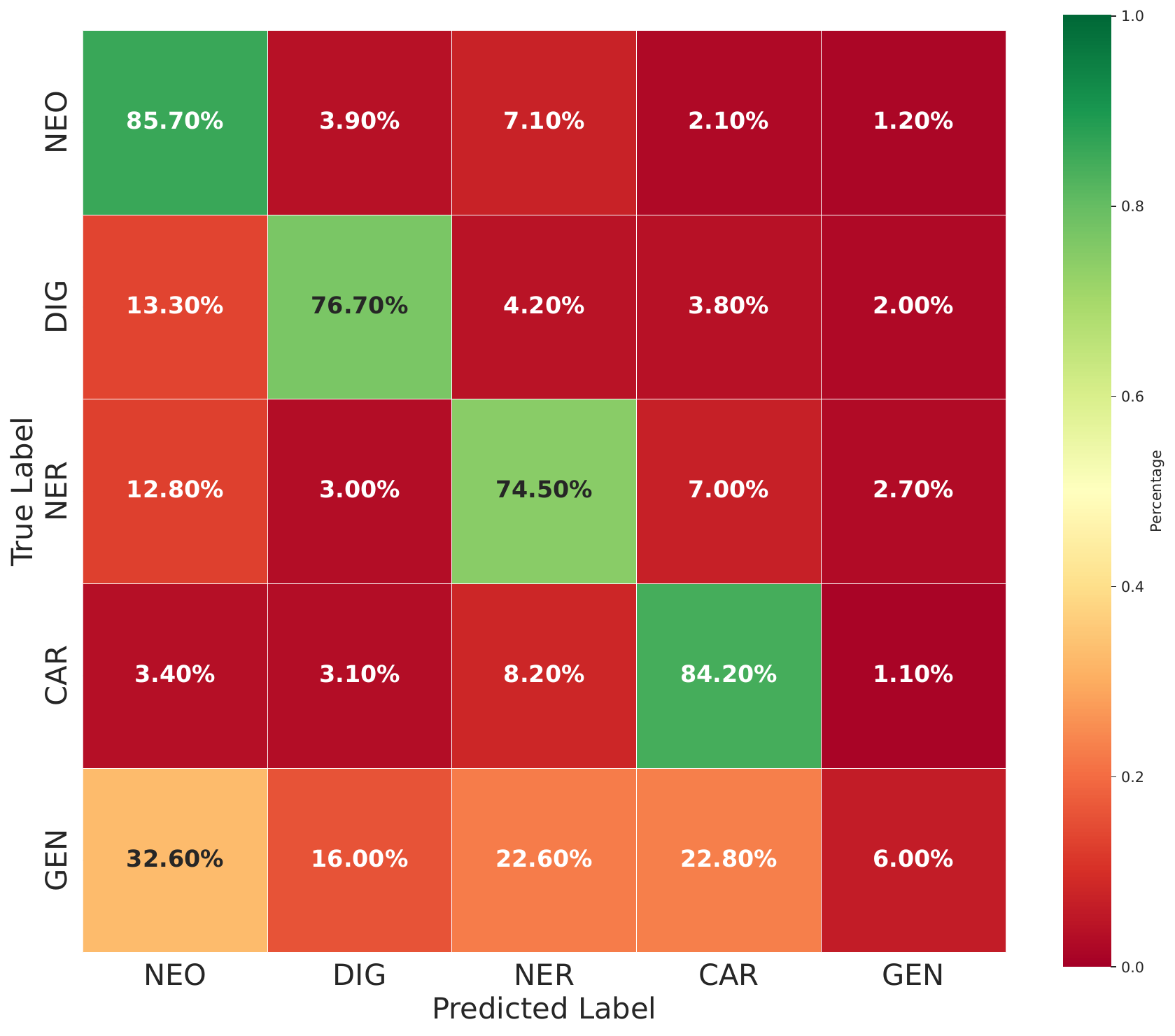}
\end{minipage}
\caption{Confusion matrices for DeBERTa-v3 over the balanced sample.Left: counts. Right: normalized percentages.\newline NEO: neoplasms; DIG: digestive; NER: nervous; CAR: cardiovascular; GEN: general.}
\label{fig:confusion}
\end{figure*}

\subsection{Predictive performance}\label{subsec:predictive_performance}

The model is evaluated over 5,000 abstracts, with 1,000 texts per randomly sampled from each category of the training split. Figure~\ref{fig:confusion} shows the confusion matrices in counts and normalized form. The diagonal is strong for the four specific categories. Neoplasms reach 85.70\% recall, cardiovascular diseases 84.20\%, digestive diseases 76.70\%, and nervous system diseases 74.50\%. The general pathological category collapses to 6\% accuracy. In the count matrix, neoplasms concentrate a large share of the errors of the other categories, and 326 of the 1,000 general cases are misclassified as neoplasms. The aggregate F1 score is 0.60.

The per-category pattern is consistent with the semantics of the target labels. The four specific categories correspond to clinical ontologies with well-characterized vocabularies: oncological processes (neoplasms), hepatobiliary and gastrointestinal pathology (digestive), central and peripheral nervous system disorders (nervous), and cardiac and vascular pathology (cardiovascular). In these cases, the entailment engine discriminates the hypotheses confidently because the premise--hypothesis compatibility is anchored to a recognizable lexical field. The semantic overlap registered in the off-diagonal cells is also clinically interpretable: digestive-to-neoplasms errors often relate to hepatic oncology cases, and cardiovascular-to-general errors reflect shared hemodynamic descriptors. The pathological general category, by contrast, has a transversal definition that includes systemic states such as fever, fatigue, weakness and diffuse inflammation, which overlap lexically with every specific organ specialty; the collapse of its recall to 6\%, with the missing mass distributed across the four specific categories, is a quantitative confirmation of this structural ambiguity and the main motivation for the dedicated error analysis protocol reported in Section~\ref{subsec:debugging_results}.

\subsection{Explainability comparison}\label{subsec:explainability_comparison}

\begin{figure*}[!t]
\centering
\includegraphics[width=0.95\textwidth]{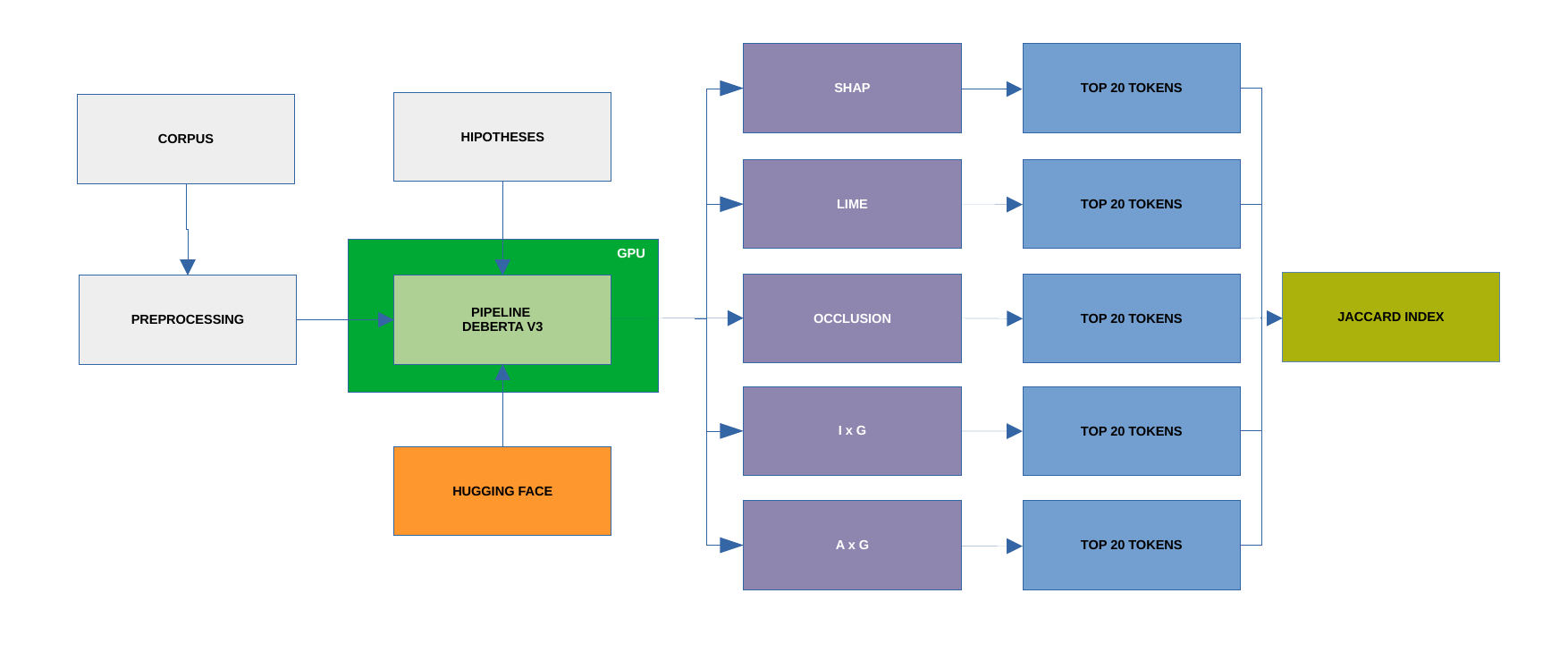}
\caption{Architecture for the comparison of explanation methods. Outputs of the five methods are standardized through Top-20 token extraction and compared with the Jaccard index.}
\label{fig:architecture_comparison}
\end{figure*}

The primary objective of this experiment is to evaluate the local consistency of post-hoc explanation methods under high-certainty predictive scenarios, thereby directly quantifying the magnitude of the disagreement problem across divergent attribution families in clinical text classification. To achieve this, the evaluation operates through the comparative pipeline depicted in Figure~\ref{fig:architecture_comparison}. 

\begin{table*}[!t]
\centering
\caption{Correctly classified cases used in the explainability comparison. Each abstract is summarized by its dominant clinical content; the full texts are available in the released experimental configuration.}
\label{tab:resumen_casos}
\scriptsize
\begin{tabularx}{\textwidth}{l X}
\toprule
Category & Summary of the abstract \\
\midrule
Neoplasms & Aggressive facial melanoma with neural invasion, successfully treated by surgical resection, radiotherapy, and reconstructive muscle flap. \\
Digestive & Optimized hepatic surgical technique using light compression bands to minimize tissue damage and hypoxia during vascular occlusion. \\
Nervous & Plasticity of the central nervous system in children (3--6 years) to coordinate oropharyngeal musculature after corrective speech surgery. \\
Cardiovascular & Long-term safety study of verapamil, highlighting its efficacy in blood pressure control and its positive effect on HDL cholesterol without altering the metabolic profile. \\
General pathological & Reinterpretation of mandibular inflammation as a muscular dysfunction caused by mechanical overload rather than infection, validated by improvement with relaxation techniques. \\
\bottomrule
\end{tabularx}
\end{table*}

The architecture channels correctly classified medical abstracts through the five selected XAI methods (SHAP, LIME, Occlusion, Input$\times$Gradient, and Attention$\times$Gradient), standardizes their outputs by isolating the Top-20 most influential tokens, and computes pairwise Jaccard agreement metrics to evaluate inter-explainer stability. To establish a controlled and reproducible empirical baseline, Table~\ref{tab:resumen_casos} summarizes the five representative, correctly classified abstracts selected for this comparative audit—one per clinical category. Each case was chosen because it contains unambiguous, domain-specific clinical descriptors that a human expert would directly associate with the true diagnostic label. This single-case per category design ($N=1$) isolates inter-method variance from inter-text variability, providing an optimal benchmark to verify whether divergent attribution algorithms converge on a shared clinical core. Tables~\ref{tab:neoplasia}--\ref{tab:general} detail the resulting pairwise Jaccard metrics and shared token sets for each diagnostic class. Across the four specific categories, the attribution methods demonstrate a clear qualitative convergence around key diagnostic entities: 

\begin{itemize}
    \item \textbf{Neoplasms (Table~\ref{tab:neoplasia}):} The token \textit{metastatic} is selected by all five methods, anchoring a compact oncological core (\textit{facial, malignant, melanoma, primary, nerve}) with a category average Jaccard index of $0.335 \pm 0.091$.
    \item \textbf{Digestive Diseases (Table~\ref{tab:digestive}):} All methods align on \textit{liver} as the primary anatomical descriptor ($0.299 \pm 0.114$), with hepatobiliary terms (\textit{hepatic, hepatocytes, parenchyma}) consistently highlighted across method pairs.
    \item \textbf{Nervous System Diseases (Table~\ref{tab:nervous}):} The structural token \textit{flap} forms the universal consensus ($0.242 \pm 0.107$), although attribution spreads across functional rehabilitation terms (\textit{speech, incompetence}).
    \item \textbf{Cardiovascular Diseases (Table~\ref{tab:cardio}):} The core centers on \textit{parameters} and \textit{metabolic} ($0.282 \pm 0.097$), with the model-agnostic pair (SHAP--LIME) reaching the highest agreement in the study ($0.529$).
\end{itemize}

\begin{table}[!t]
\centering
\caption{Pairwise Jaccard values for a correctly classified neoplasm case.}
\label{tab:neoplasia}
\scriptsize
\begin{tabular}{lc}
\toprule
Comparison & Jaccard \\
\midrule
SHAP vs.\ LIME & 0.333 \\
SHAP vs.\ IxG & 0.300 \\
SHAP vs.\ Occlusion & 0.500 \\
SHAP vs.\ AxG & 0.333 \\
LIME vs.\ IxG & 0.185 \\
LIME vs.\ Occlusion & 0.200 \\
LIME vs.\ AxG & 0.300 \\
IxG vs.\ Occlusion & 0.368 \\
IxG vs.\ AxG & 0.400 \\
Occlusion vs.\ AxG & 0.428 \\
\midrule
Global & 0.335 $\pm$ 0.091 \\
\bottomrule
\end{tabular}
\end{table}

\begin{table}[!t]
\centering
\caption{Pairwise Jaccard values for a correctly classified digestive case.}
\label{tab:digestive}
\scriptsize
\begin{tabular}{lc}
\toprule
Comparison & Jaccard \\
\midrule
SHAP vs.\ LIME & 0.352 \\
SHAP vs.\ IxG & 0.200 \\
SHAP vs.\ Occlusion & 0.111 \\
SHAP vs.\ AxG & 0.157 \\
LIME vs.\ IxG & 0.318 \\
LIME vs.\ Occlusion & 0.315 \\
LIME vs.\ AxG & 0.421 \\
IxG vs.\ Occlusion & 0.238 \\
IxG vs.\ AxG & 0.473 \\
Occlusion vs.\ AxG & 0.411 \\
\midrule
Global & 0.299 $\pm$ 0.114 \\
\bottomrule
\end{tabular}
\end{table}

\begin{table}[!t]
\centering
\caption{Pairwise Jaccard values for a correctly classified nervous system case.}
\label{tab:nervous}
\scriptsize
\begin{tabular}{lc}
\toprule
Comparison & Jaccard \\
\midrule
SHAP vs.\ LIME & 0.200 \\
SHAP vs.\ IxG & 0.368 \\
SHAP vs.\ Occlusion & 0.352 \\
SHAP vs.\ AxG & 0.111 \\
LIME vs.\ IxG & 0.333 \\
LIME vs.\ Occlusion & 0.380 \\
LIME vs.\ AxG & 0.130 \\
IxG vs.\ Occlusion & 0.291 \\
IxG vs.\ AxG & 0.120 \\
Occlusion vs.\ AxG & 0.136 \\
\midrule
Global & 0.242 $\pm$ 0.107 \\
\bottomrule
\end{tabular}
\end{table}

\begin{table}[!t]
\centering
\caption{Pairwise Jaccard values for a correctly classified cardiovascular case.}
\label{tab:cardio}
\scriptsize
\begin{tabular}{lc}
\toprule
Comparison & Jaccard \\
\midrule
SHAP vs.\ LIME & 0.529 \\
SHAP vs.\ IxG & 0.217 \\
SHAP vs.\ Occlusion & 0.250 \\
SHAP vs.\ AxG & 0.150 \\
LIME vs.\ IxG & 0.307 \\
LIME vs.\ Occlusion & 0.238 \\
LIME vs.\ AxG & 0.260 \\
IxG vs.\ Occlusion & 0.272 \\
IxG vs.\ AxG & 0.240 \\
Occlusion vs.\ AxG & 0.353 \\
\midrule
Global & 0.282 $\pm$ 0.097 \\
\bottomrule
\end{tabular}
\end{table}

In sharp contrast, when inspecting the General pathological category (Table~\ref{tab:general}), no single token is shared across all five methods, and the mean agreement drops to its lowest value ($0.171 \pm 0.088$, reaching a minimum pair agreement between LIME and Occlusion of only $0.036$). These per-category results confirm that explanatory agreement is strongly conditioned by clinical specificity: unambiguous diagnostic labels foster a stable explanation core across method families, whereas broad, semantically overlapping classes lead to fragmented attributions.

\begin{table}[!t]
\centering
\caption{Pairwise Jaccard values for a general pathological case. No token is shared by all five methods.}
\label{tab:general}
\scriptsize
\begin{tabular}{lc}
\toprule
Comparison & Jaccard \\
\midrule
SHAP vs.\ LIME & 0.136 \\
SHAP vs.\ IxG & 0.227 \\
SHAP vs.\ Occlusion & 0.059 \\
SHAP vs.\ AxG & 0.188 \\
LIME vs.\ IxG & 0.267 \\
LIME vs.\ Occlusion & 0.036 \\
LIME vs.\ AxG & 0.153 \\
IxG vs.\ Occlusion & 0.291 \\
IxG vs.\ AxG & 0.280 \\
Occlusion vs.\ AxG & 0.095 \\
\midrule
Global & 0.171 $\pm$ 0.088 \\
\bottomrule
\end{tabular}
\end{table}

\subsection{Cross-class agreement analysis}\label{subsec:cross_class}

Explanatory agreement is strongly conditioned by the underlying mathematical formulation of the attribution method rather than being a static property of the language model (Table~\ref{tab:cross_class}). The highest convergence is consistently achieved between methods belonging to the same algorithmic family: model-agnostic perturbation techniques (SHAP--LIME, mean $0.31 \pm 0.14$) and internal model-flow approaches (Input~$\times$ ~ Gradient--Attention~$\times$ ~ Gradient, mean $0.30 \pm 0.12$). In contrast, cross-family comparisons involving attention-based mechanisms yield the lowest agreement in the study (e.g. SHAP--Attention~$\times$ ~ Gradient, mean $0.19 \pm 0.08$). This divergence empirically confirms the disagreement problem in clinical NLP, demonstrating that a single explainer provides an incomplete view of the model's decision boundary.

\begin{table*}[!t]
\centering
\caption{Pairwise Jaccard index per category and method pair. NEO: neoplasms; DIG: digestive; NER: nervous; CAR: cardiovascular; GEN: general.}
\label{tab:cross_class}
\scriptsize
\begin{tabular}{lccccc|c}
\toprule
Comparison & NEO & DIG & NER & CAR & GEN & Mean $\pm$ SD \\
\midrule
SHAP vs.\ LIME & 0.333 & 0.352 & 0.200 & 0.529 & 0.136 & 0.31 $\pm$ 0.14 \\
SHAP vs.\ IxG & 0.300 & 0.200 & 0.368 & 0.217 & 0.227 & 0.26 $\pm$ 0.06 \\
SHAP vs.\ Occlusion & 0.500 & 0.111 & 0.352 & 0.250 & 0.059 & 0.25 $\pm$ 0.16 \\
SHAP vs.\ AxG & 0.333 & 0.157 & 0.111 & 0.150 & 0.188 & 0.19 $\pm$ 0.08 \\
LIME vs.\ IxG & 0.185 & 0.318 & 0.333 & 0.307 & 0.267 & 0.28 $\pm$ 0.05 \\
LIME vs.\ Occlusion & 0.200 & 0.315 & 0.380 & 0.238 & 0.036 & 0.23 $\pm$ 0.12 \\
LIME vs.\ AxG & 0.300 & 0.421 & 0.130 & 0.260 & 0.153 & 0.25 $\pm$ 0.11 \\
IxG vs.\ Occlusion & 0.368 & 0.238 & 0.291 & 0.272 & 0.291 & 0.29 $\pm$ 0.04 \\
IxG vs.\ AxG & 0.400 & 0.473 & 0.120 & 0.240 & 0.280 & 0.30 $\pm$ 0.12 \\
Occlusion vs.\ AxG & 0.428 & 0.411 & 0.136 & 0.353 & 0.095 & 0.28 $\pm$ 0.14 \\
\midrule
Global & 0.33 $\pm$ 0.09 & 0.30 $\pm$ 0.11 & 0.24 $\pm$ 0.11 & 0.28 $\pm$ 0.10 & 0.17 $\pm$ 0.09 & 0.27 $\pm$ 0.06 \\
\bottomrule
\end{tabular}
\end{table*}

Furthermore, explanation stability exhibits a direct positive correlation with predictive confidence across clinical domains (Figure~\ref{fig:confusion} and Table~\ref{tab:cross_class}). Unambiguous diagnostic categories with well-defined clinical vocabularies—such as Neoplasms ($85.70\%$ recall, mean Jaccard $0.33 \pm 0.09$) and Cardiovascular diseases ($84.20\%$ recall, mean Jaccard $0.28 \pm 0.10$)—foster a stable explanatory core centered on key pathological descriptors (\textit{metastatic}, \textit{liver}, \textit{hypertension}). In contrast, diagnostic ambiguity in the General pathological category ($6.00\%$ recall) induces a severe collapse in explanatory convergence (mean Jaccard $0.17 \pm 0.09$, reaching a minimum of $0.036$ between LIME and Occlusion). When the model lacks a dominant clinical feature, attribution mass disperses onto secondary grammatical particles, rendering the post-hoc explanation unstable.

% In the specific categories, the shared tokens correspond to clinical descriptors such as \textit{metastatic}, \textit{liver}, \textit{flap}, \textit{hypertension}, or \textit{cholesterol}. In the general category, the shared tokens are limited to nonspecific terms such as \textit{unknown}, \textit{diffuse}, or \textit{mandible}. The absence of a shared clinical core is consistent with the low accuracy observed for that class.

\subsection{Error analysis the general pathological category}\label{subsec:debugging_results}

\begin{figure*}[!t]
\centering
\includegraphics[width=0.95\textwidth]{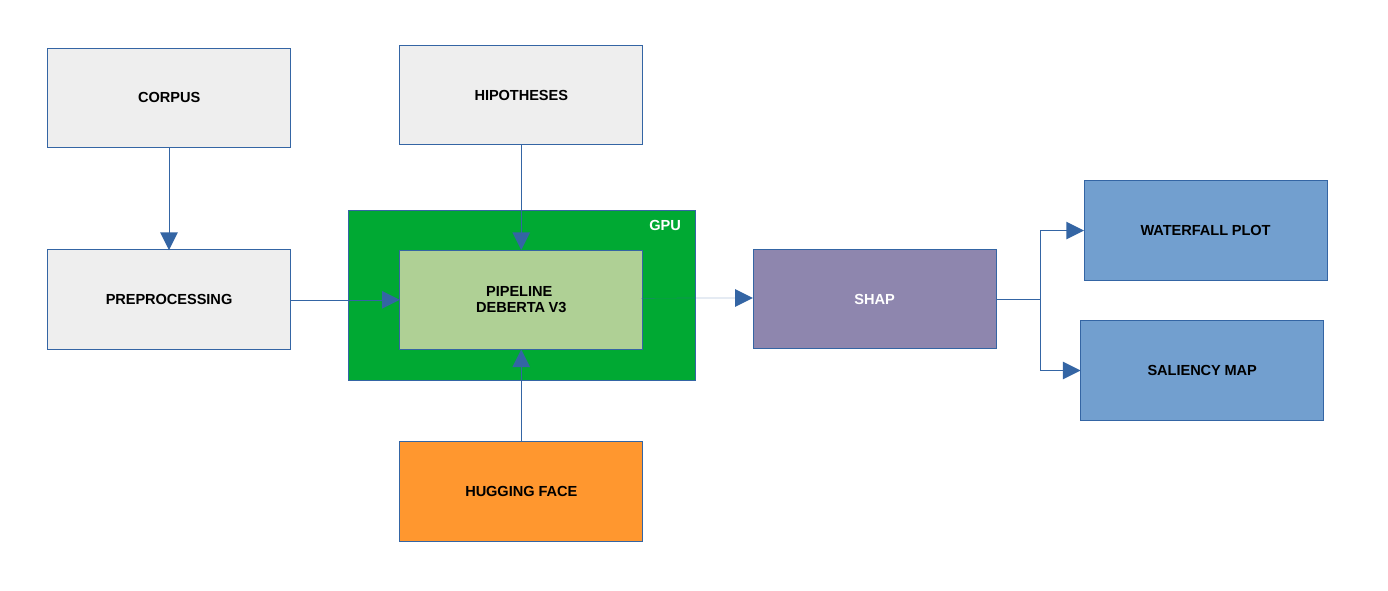}
\caption{Architecture for the error analysis protocol of the general pathological category. SHAP attributions are visualized through saliency maps and waterfall plots.}
\label{fig:architecture_debugging}
\end{figure*}

Figure~\ref{fig:architecture_debugging} shows the architecture used for the error analysis protocol. The pipeline is kept until the prediction stage, and SHAP is then used to inspect the decision of the model over the general pathological cases that are misclassified.

\begin{table*}[!t]
\centering
\caption{Summary of the five misclassified general pathological cases inspected with SHAP. \emph{Predicted}: erroneous category assigned by DeBERTa-v3. \emph{Top driving tokens}: tokens with the largest positive SHAP attribution per case. \emph{Failure pattern}: LH lexical hypersensitivity, SO semantic overlap, LAC loss of attribution coherence (see text).}
\label{tab:debug_summary}
\scriptsize
\begin{tabularx}{\textwidth}{c l X l}
\toprule
Case & Predicted & Top driving tokens & Failure pattern \\
\midrule
1 & Cardiovascular & \textit{hemodynamic}, \textit{atherosclerosis}, \textit{glomerular} & SO \\
2 & Neoplasms & \textit{malignancy}, \textit{tumor}, \textit{carcinoma} & LH \\
3 & Cardiovascular & \textit{pressure}, \textit{blood} & SO \\
4 & Neoplasms & \textit{it}, \textit{diagnose} (no robust descriptor) & LAC \\
5 & Neoplasms & \textit{biopsy} & LH \\
\bottomrule
\end{tabularx}
\end{table*} 

Five misclassified general cases are inspected. Each case is summarized in Table~\ref{tab:debug_summary} together with the dominant tokens identified by SHAP, the predicted category, and the failure pattern assigned during the qualitative analysis. Table~\ref{tab:analisis_textos_nuevos} complements this summary with a clinical description of the content of each misclassified abstract, which makes the failure patterns concrete: the texts refer to systemic processes (renal, infectious, procedural) that the model assigns to a specific specialty because of a dominant lexical cue. Two representative cases are described in detail below; the remaining three are discussed in compact form at the end of this subsection.

\begin{table*}[!t]
\centering
\caption{Clinical content of the five misclassified general pathological cases. \emph{Predicted}: erroneous category assigned by DeBERTa-v3. The true category in all cases is \emph{General pathological conditions}.}
\label{tab:analisis_textos_nuevos}
\scriptsize
\begin{tabularx}{\textwidth}{c l X}
\toprule
Case & Predicted & Summary of the abstract \\
\midrule
1 & Cardiovascular & Pathophysiology of glomerular damage and scarring in chronic kidney disease, highlighting the interaction between genetic and metabolic factors. \\
2 & Neoplasms & Experimental study on the malignant transformation of keratinocytes infected with HPV-18, linking tumour progression to chromosomal instability and cell-cycle failures. \\
3 & Cardiovascular & Investigates platelet dysfunction in uraemic patients, identifying excess nitric oxide as the key mediator of bleeding and its therapeutic reversibility. \\
4 & Neoplasms & Studies streptococcal pharyngitis, its treatment with penicillin, and the systemic risks derived from inadequate management of the bacterial infection. \\
5 & Neoplasms & Safety evaluation of outpatient lung biopsy in 169 patients, showing a low incidence of severe complications and the feasibility of same-day discharge. \\
\bottomrule
\end{tabularx}
\end{table*}

\begin{figure*}[!t]
\centering
\begin{minipage}{0.49\textwidth}
\centering
\includegraphics[width=\linewidth]{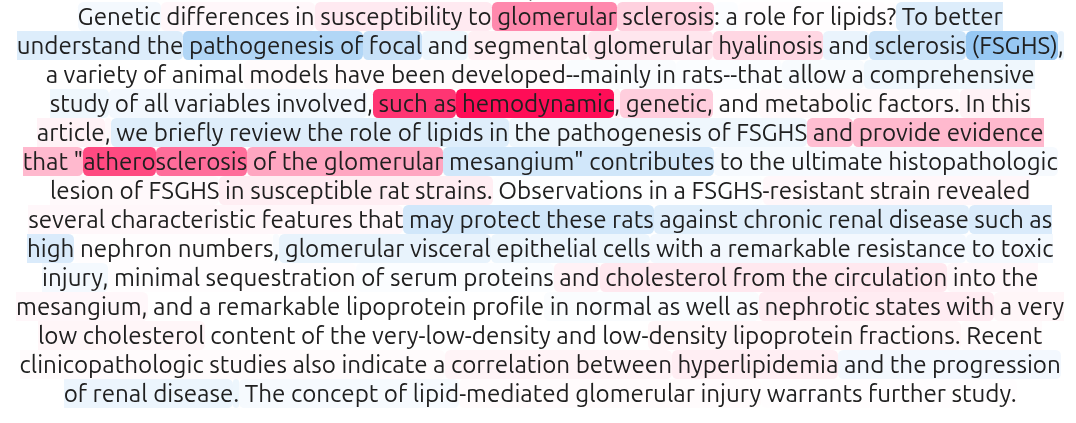}
\end{minipage}\hfill
\begin{minipage}{0.49\textwidth}
\centering
\includegraphics[width=\linewidth]{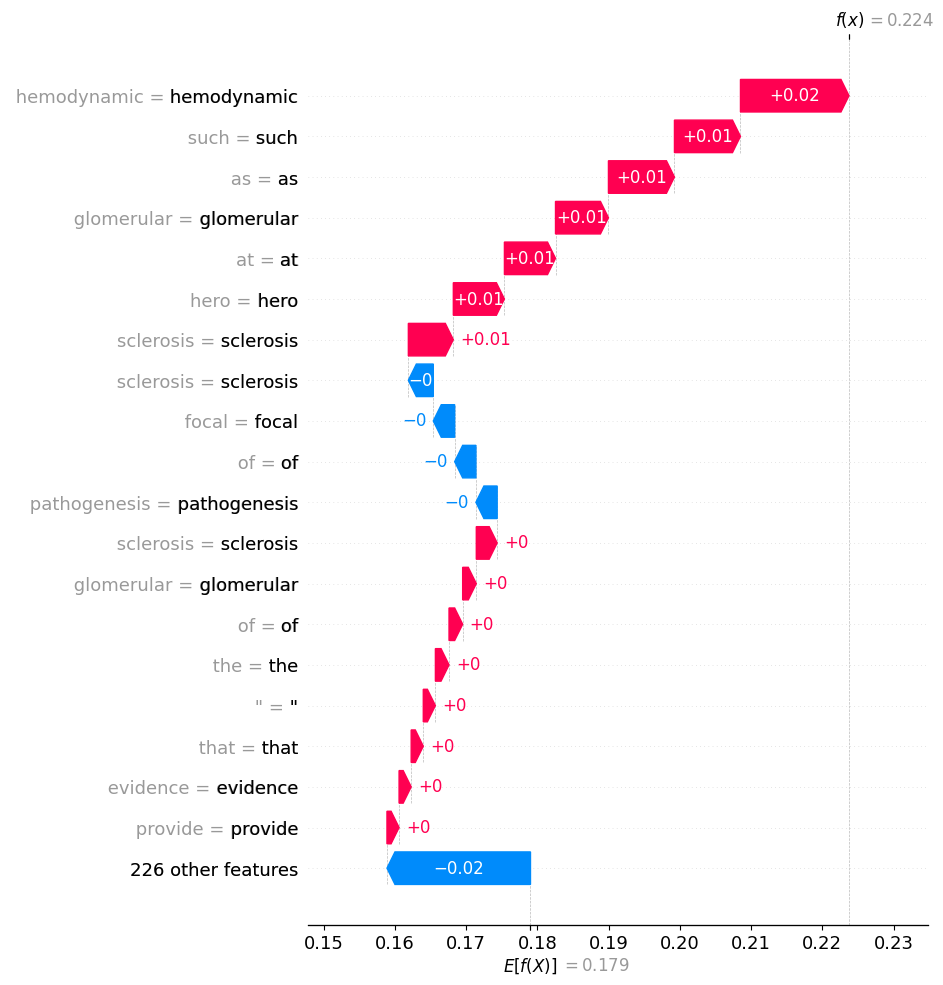}
\end{minipage}
\caption{Error analysis of a general case misclassified as cardiovascular. Left: saliency map. Right: SHAP waterfall plot.}
\label{fig:debug_text1}
\end{figure*}

The first case is classified as cardiovascular. The saliency map and the waterfall plot, shown in Figure~\ref{fig:debug_text1}, indicate that tokens with vascular load such as \textit{hemodynamic}, \textit{atherosclerosis}, and \textit{glomerular} push the prediction toward the cardiovascular category. The clinical acronym FSGHS, which refers to focal segmental glomerulosclerosis, has an attribution below the level needed to correct the prediction. The model thus prioritizes shared vascular descriptors over the systemic context of the abstract, illustrating the semantic overlap pattern: descriptors that are anatomically vascular inherit the diagnostic weight of a neighbouring specific category even when the abstract describes a systemic renal process.

\begin{figure*}[!t]
\centering
\begin{minipage}{0.49\textwidth}
\centering
\includegraphics[width=\linewidth]{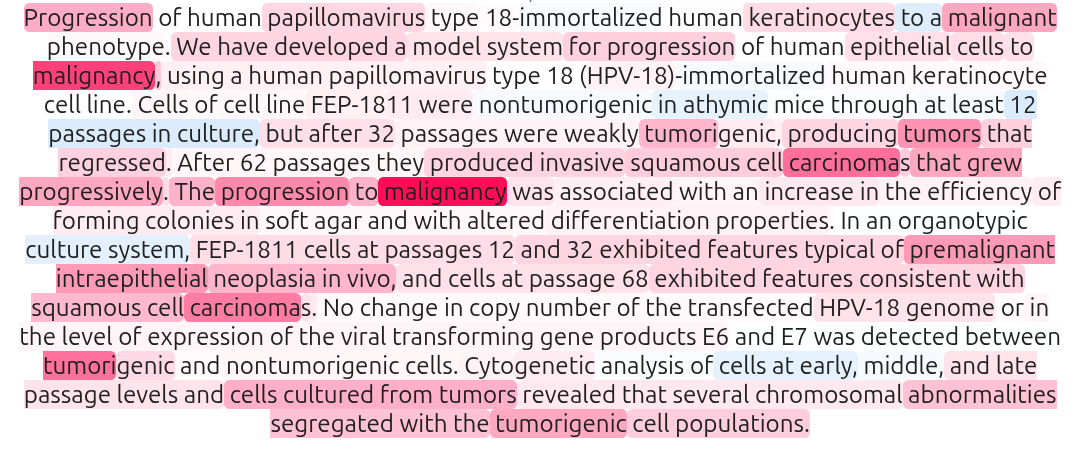}
\end{minipage}\hfill
\begin{minipage}{0.49\textwidth}
\centering
\includegraphics[width=\linewidth]{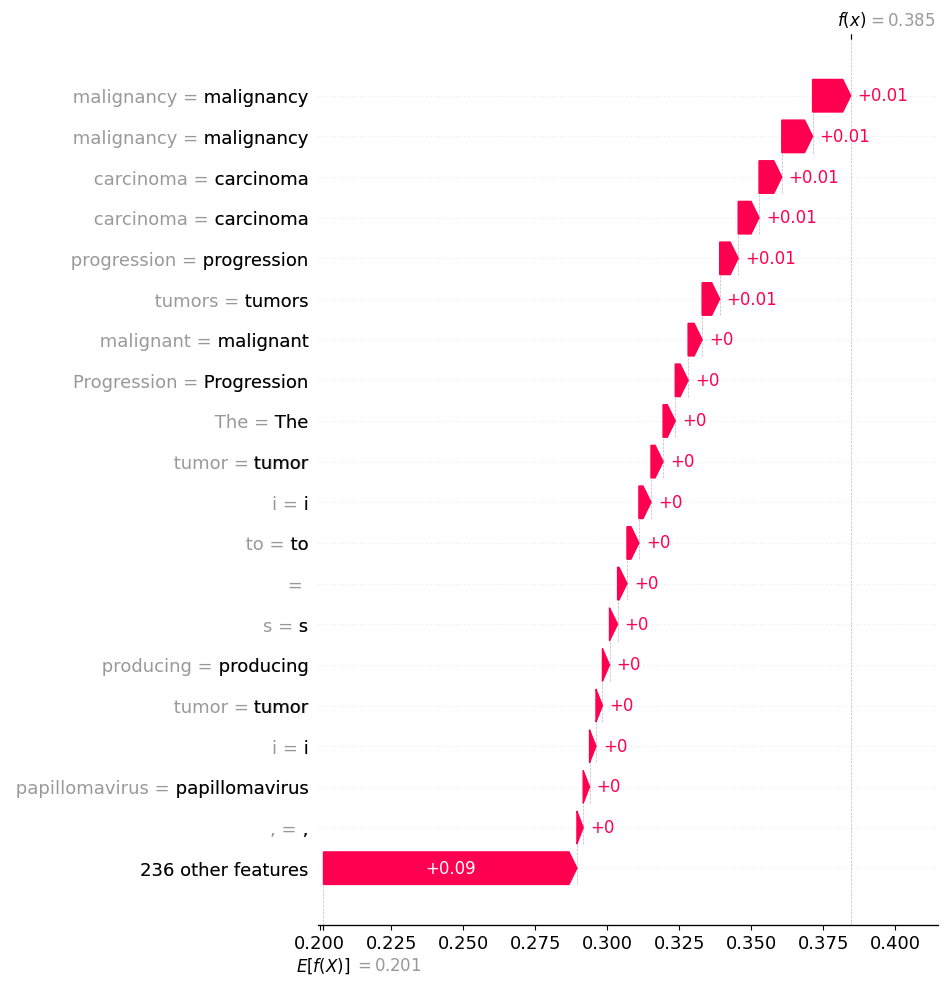}
\end{minipage}
\caption{Error analysis of a general case misclassified as neoplasm. Left: saliency map. Right: SHAP waterfall plot.}
\label{fig:debug_text2}
\end{figure*}

The second case is classified as neoplasms. Figure~\ref{fig:debug_text2} shows that \textit{malignancy}, \textit{tumor}, and \textit{carcinoma} are the main drivers of the prediction. The model identifies the correct label, but the explanation relies almost exclusively on explicit oncological terms and ignores the cellular transformation mediated by HPV-18, which is the central process of the abstract. This illustrates the lexical hypersensitivity pattern: the prediction is correct only because the abstract foregrounds explicit oncological terminology, but the underlying reasoning is fragile for texts that describe oncological processes without direct diagnostic vocabulary.

The remaining three cases corroborate the same patterns. The third case describes coagulation problems in patients with renal failure and is misclassified as cardiovascular because \textit{pressure} and \textit{blood} capture most of the predictive weight; negative mentions of renal mass are not processed as evidence against the circulatory class, replicating the semantic overlap pattern of case~1. The fourth case corresponds to an attribution collapse in which neither the saliency map nor the waterfall plot reveals a coherent clinical pattern, with the importance mass drifting towards grammatical particles such as the pronoun \textit{it} or the verb \textit{diagnose}; this is the loss of attribution coherence pattern. The fifth case is misclassified as neoplasms because the token \textit{biopsy} alone dominates the prediction, even though the abstract refers to an outpatient procedure with no oncological context; this confirms the lexical hypersensitivity pattern observed in case~2.

Across the five inspected cases, three recurring failure patterns are therefore identified. First, \emph{lexical hypersensitivity} (cases~2 and~5): a single token such as \textit{biopsy} or \textit{malignancy} can dominate the prediction and shift the case toward neoplasms. Second, \emph{semantic overlap} (cases~1 and~3): shared descriptors such as vascular or renal terms lead to a specific category when the abstract has a systemic nature, and negative evidence is not assimilated. Third, \emph{loss of attribution coherence} (case~4): when no robust clinical descriptor is available, the attribution mass is distributed over grammatical particles or nonspecific terms, producing an erratic explanation rather than a structured clinical justification. This last pattern is consistent with the low Jaccard values observed for the general category in Table~\ref{tab:cross_class}, and it suggests that the health of the explanation itself can be used as a diagnostic signal for the reliability of the underlying prediction.

Although this audit focuses on five structured cases, these instances are not isolated anomalies. Instead, they constitute a representative qualitative taxonomy that characterizes three systemic failure mechanisms inherent to zero-shot NLI in transformer architectures:
\begin{enumerate}
    \item Lexical Hypersensitivity: Over-reliance on explicit high-salience terms (e.g., \textit{biopsy}, \textit{malignancy}) that trigger premature category assignment regardless of procedural or diagnostic context.
    \item Semantic Overlap: Inability to weigh negative clinical evidence when shared physiological descriptors (e.g., vascular or renal terms) overlap with specific organ specialties.
    \item Loss of Attribution Coherence: Stochastic dispersion of feature importance onto secondary grammatical particles when the input text lacks a dominant clinical feature.
\end{enumerate}

\section{Discussion}\label{sec:discussion}

The results show two connected findings. The first is that explanation agreement is not a stable property of the model alone. It depends on the category and on the family of the explanation method. Agreement is higher within model-agnostic methods (SHAP--LIME, 0.31 on average) and within methods based on internal model flow (Input x Gradient--Attention x Gradient, 0.30), and lower between agnostic and attention-based methods (SHAP--Attention x Gradient, 0.19). This pattern confirms the disagreement problem in a clinical text setting and indicates that a single explainer can give a partial view of the decision process, in line with the practitioner-level concerns reported by \citet{Krishna2022} and the broader critique of \citet{Longo2024}.

The second finding is the relation between predictive performance and explanation stability. The four specific categories combine higher accuracy with a shared clinical core in the explanations, with the per-category Jaccard average spanning 0.24 (nervous system) to 0.33 (neoplasms). The general category combines the lowest accuracy (6\%), the absence of a shared clinical core, and the lowest Jaccard value (0.17). When the model has a robust clinical descriptor, the explanation methods tend to converge on that descriptor; for neoplasms, the token \textit{metastatic} is selected by all five methods, and for digestive diseases, \textit{liver} plays the same role. When the input is semantically ambiguous, the attributions become fragmented and the disagreement increases. This observation supports the combined use of several explanation methods and a quantitative agreement metric instead of a single visual explanation, complementing the diagnostic study of \citet{Atanasova2020}, who reported method-dependent highlighting for text classifiers but did not quantify the divergence per category.

The error analysis provides additional evidence. The five misclassified general cases summarized in Table~\ref{tab:debug_summary} do not fail because of random noise only. They fail because the model assigns a large weight to specific lexical cues that belong to adjacent categories: a single token such as \textit{biopsy} (case~5) flips the prediction towards neoplasms even when the abstract has no oncological context, and shared vascular descriptors pull systemic cases towards cardiovascular (cases~1 and~3). Most tellingly, case~4 exhibits an attribution collapse in which the importance mass drifts onto grammatical particles such as the pronoun \textit{it} or the verb \textit{diagnose}, mirroring the loss of attribution coherence reported in the qualitative literature on explanation failure \citep{Jacovi2020}. This behavior is problematic in a zero-shot setting, where no task-specific training corrects the prior induced by the pretraining corpus, and where the absence of supervision means that the only regulator of the decision boundary is the natural language inference scoring mechanism described in Section~\ref{subsec:inference}.

A practical implication follows from this analysis: broad diagnostic labels such as \emph{general pathological conditions} are harder to audit than specific ontologies because their language structurally overlaps with several specific categories. The transformer, lacking task-specific supervision, inherits the lexical statistics of its pretraining corpus and amplifies whatever diagnostic cue is most strongly associated with a specific class. From an accountability standpoint this makes the general category an unsafe target for an automated decision system, because the post-hoc explanation cannot disambiguate a legitimate systemic diagnosis from a positive mention of an adjacent specialty. This motivates the design recommendation, also echoed by \citet{EDPS2023} and the regulatory debate around the European AI Act \citep{AIAct2021, DiMarino2025}, that clinical classification targets should favour specific ontologies whenever auditability is a requirement, and that the residual systemic class should be either refined through a secondary multi-label structure or routed to manual review.

The findings are aligned with previous reports on the disagreement problem \citep{Longo2024, Krishna2022} and with the recommendation to evaluate explanations quantitatively through standardized benchmarks \citep{Sithakoul2024, Hedstrom2023, Atanasova2020}. They also complement related XAI work for text classification in adjacent domains \citep{Jang2023} by showing, specifically for a transformer-based medical zero-shot setting, how explanation agreement can be used as a diagnostic signal of model uncertainty rather than only as a property to be reported. Two aspects of this signal deserve emphasis. First, the Jaccard index used here measures set overlap without considering ranking; a complementary study using rank-aware measures such as the Rank-Biased Overlap \citep{AlvarezMelis2018} or fidelity metrics such as AOPC-sufficiency and comprehensiveness \citep{Hedstrom2023} would help disentangle cases in which the same tokens are selected but ranked very differently. Second, the present protocol samples one case per category for the comparative analysis; while this is sufficient to expose the divergence pattern across method families, an inter-text variability study with a larger $N$ is needed before extrapolating the per-category averages to a population of inputs. Both extensions are discussed in Section~\ref{sec:limitations}.

\section{Scope of Validity and Limitations}\label{sec:limitations}

The findings and empirical conclusions of this study are bounded by specific experimental conditions and methodological design choices, which define its scope of validity:

\emph{Construct Validity and Metric Scope:} The pairwise Jaccard index over Top-20 token sets captures set overlap while discarding token ranking and absolute attribution magnitude. Selecting $k=20$ tokens strikes an optimal trade-off between capturing dominant clinical entities and filtering out statistical noise; however, $k$ remains a design parameter. Complementary ranking-aware measures (e.g., Rank-Biased Overlap) and causal fidelity tests (e.g., AOPC-sufficiency and comprehensiveness) represent natural extensions to evaluate fine-grained token hierarchies.

\emph{Internal Validity and Sample Protocol:} The comparative XAI phase employs a controlled single-case per category design ($N=1$) to isolate inter-method variance from inter-text variability, establishing a reproducible baseline across explainer families. Consequently, the reported standard deviations reflect variability across method pairs rather than across inputs. Similarly, the error audit examines five structured cases to construct a qualitative failure taxonomy. While these samples successfully expose systemic divergence patterns, broader population-level sampling will further quantify the exact prevalence of each failure mode across large patient cohorts.

\emph{External Validity and Model Domain:} The evaluation focuses on zero-shot inference using DeBERTa-v3-large on the Medical Abstracts corpus. This setup audits the native representational space of the transformer without task-specific fine-tuning. Replicating this protocol across alternative backbones (e.g., BioBERT, ClinicalBERT) or supervised fine-tuning regimes will delineate how domain-specific adaptation stabilizes explanation agreement. Advanced frameworks such as LLiMe and AttnLRP were excluded due to computational constraints and compatibility limits with DeBERTa-v3, representing key targets for future integration.

\emph{Reliability and Stochastic Stability:} Stochastic components (sampling, perturbation-based explainers) are governed by a fixed random seed, and SHAP/LIME attributions are stabilized using a 1,000 perturbation budget per sample. While this budget keeps estimator variance within a narrow band, evaluating empirical variance across multiple random seeds remains a valuable step for future benchmark extensions.

\section{Conclusions and Future Work}\label{sec:conclusions}

This paper presents a comparative explainability framework for DeBERTa-v3 in zero-shot medical abstract classification. Five explanation methods spanning three algorithmic families were evaluated through Top-20 token attribution and the pairwise Jaccard index over the Medical Abstracts corpus under a natural language inference engine with five enriched hypotheses per category.

The empirical findings demonstrate that explanatory agreement directly tracks predictive performance across clinical domains: specific categories with well defined ontologies foster high accuracy and a stable, shared clinical core across explainers, whereas broad, semantically overlapping classes induce predictive collapse and severe attribution fragmentation. Furthermore, the error analysis protocol identified lexical hypersensitivity, semantic overlap, and loss of attribution coherence as three systemic failure mechanisms in zero-shot NLI architectures.

Three practical indications follow from this audit. First, single-explainer audits are insufficient for transformer-based medical classifiers because attribution paths depend heavily on the underlying explainer family. Second, pairwise agreement metrics over standardized $Top\text{-}k$ tokens provide a compact, actionable diagnostic signal of explanation reliability. Third, target taxonomies in clinical NLP should prioritize specific ontologies over broad diagnostic labels whenever auditability is a requirement.

Building on these findings, future research will advance along four strategic avenues. First, the evaluation benchmark will be expanded to incorporate cutting-edge attribution methods—such as generative neighborhood explainers (\textit{LLiMe}), layer-wise relevance propagation (\textit{AttnLRP}), and context-aware layer-wise integrated gradients (\textit{CA-LIG})—to further mitigate grammatical perturbation noise and capture deep representation flows. Second, to better accommodate explanatory uncertainty, the validation protocol will transition from crisp set-overlap metrics toward \textit{Fuzzy Jaccard} similarity measures and fuzzy set theory, enabling a more continuous and nuanced modeling of token relevance. Third, formal domain ontologies will be integrated into the standardization protocol to map raw tokens onto standardized clinical concepts, evaluating explanatory convergence at a conceptual level rather than strictly at the surface lexical level. Finally, the comparative auditing framework will be generalized and standardized beyond clinical text classification to encompass alternative domains and data modalities, including Computer Vision and tabular data, establishing a unified, domain-agnostic protocol for multi-explainer reliability.

\section*{Data availability}\label{sec:data_availability}

The Medical Abstracts corpus is described in \citet{Schopf2022} and was used under the conditions of its original release. The code and the experimental configurations used in this work are available from the authors upon request.

\section*{Declaration of competing interests}\label{sec:competing_interests}
The authors declare that they have no known competing financial interests or personal relationships that could have appeared to influence the work reported in this paper.

\section*{Use of generative AI}\label{sec:genai}
During the preparation of this work the authors used an AI assistant for code refactoring, LaTeX editing and language polishing. After using this tool, the authors reviewed and edited the content as needed and take full responsibility for the content of the publication.

\ifanonym\else
\section*{Acknowledgments}\label{sec:acknowledgments}
This work was partially funded by the RecaIA project (Ref: PID2025-\allowbreak 167834OB-\allowbreak I00) by MCIN\slash AEI\slash 10.13039\slash 501100011033 and by ``ERDF: A way of making Europe'', by the ``European Union''. This work was also supported by grant PID2025-\allowbreak 168152NB-\allowbreak C32 funded by MCIN\slash AEI\slash 10.13039\slash 501100011033, by ERDF A Way of Making Europe.
% ===== TODO: fill in before submission (only in identified version) =====
% e.g., "The authors thank Raquel Mart\'inez Espa\~na and Jos\'e Manuel Juarez
%  (University of Murcia) for their direction of the underlying
% Master's thesis. This work was supported by [grant/program] (optional)."
% ============================================================
\fi

\printcredits

\bibliographystyle{cas-model2-names}
\bibliography{bibliography/main}

\end{document}